\pdfoutput=1

\documentclass[11pt]{article}

\usepackage[table]{xcolor}
\usepackage{acl}

\usepackage{times}
\usepackage{latexsym}

\usepackage[T1]{fontenc}

\usepackage[utf8]{inputenc}

\usepackage{microtype}

\usepackage{inconsolata}

\usepackage{graphicx}
\usepackage{booktabs}
\usepackage{amsmath}
\usepackage{amssymb}
\usepackage{algorithm}
\usepackage{algorithmic}
\usepackage[capitalize,noabbrev]{cleveref}

\title{ASPO: Adaptive Sentence-Level Preference Optimization for Fine-Grained Multimodal Reasoning}

\author{
    \textbf{Yeyuan Wang\textsuperscript{1,\thanks{Equal Contribution.}}},
    \textbf{Dehong Gao\textsuperscript{2,3,\footnotemark[1]}},    \textbf{Rujiao Long\textsuperscript{4}},
    \textbf{Lei Yi\textsuperscript{4}}, \\
    \textbf{Linbo Jin\textsuperscript{4}},
    \textbf{Libin Yang\textsuperscript{2},\thanks{Corresponding Authors.}},
    \textbf{Xiaoyan Cai\textsuperscript{1},\footnotemark[2]}
    \\
    \textsuperscript{1}{Northwestern Polytechnical University, School of Automation, Xi'an, China} \\
    \textsuperscript{2}{Northwestern Polytechnical University, School of Cybersecurity, Xi'an, China}\\
    \textsuperscript{3}{Binjiang Institute of Artificial Intelligence, ZJUT, Hangzhou, China} \\
    \textsuperscript{4}{Alibaba Group, Hangzhou, China}
    \\
    \small{
    \{wangyeyuan, dehong.gdh, libiny, xiaoyanc\}@nwpu.edu.cn
    }\\
    \small{
    \{rujiao.lrj, yilei.yi, yuyi.jlb\}@alibaba-inc.com
    }
}

\begin{document}
\maketitle
\begin{abstract}
Direct Preference Optimization (DPO) has gained significant attention for its simplicity and computational efficiency in aligning large language models (LLMs). 
Recent advancements have extended DPO to multimodal scenarios, achieving strong performance. 
However, traditional DPO relies on binary preference optimization, rewarding or penalizing entire responses without considering fine-grained segment correctness, leading to suboptimal solutions.
The root of this issue lies in the absence of fine-grained supervision during the optimization process.
To address this, we propose Adaptive Sentence-level Preference Optimization (ASPO), which evaluates individual sentences for more precise preference optimization. 
By dynamically calculating adaptive rewards at the sentence level based on model predictions, ASPO enhances response content assessment without additional models or parameters. 
This significantly improves the alignment of multimodal features.
Extensive experiments show that ASPO substantially enhances the overall performance of multimodal models.
\end{abstract}

\section{Introduction}
Large Language Models (LLMs) have demonstrated remarkable capabilities in Natural Language Processing (NLP)~\cite{touvron2023llama, zhang2022opt, achiam2023gpt4}, which greatly accelerates the development of Multimodal Large Language Models (MLLMs)~\cite{kirillov2023sam, li2024llava-med, anil2023gemini}.
The training of MLLMs typically involves a pre-training stage followed by the Supervised Fine-Tuning (SFT) stage \cite{liu2024llava, instructblip}.
SFT enhances the model's ability to understand and execute multimodal instructions~\cite{liu2024llava1.5, yu2023reformulating, ye2024mplug}. 
Models trained in this manner exhibit impressive multimodal conversational abilities~\cite{yu2023reformulating, huang2023kosmos}.



\begin{figure*}[ht]
\begin{center}
\centerline{\includegraphics[width=\textwidth]{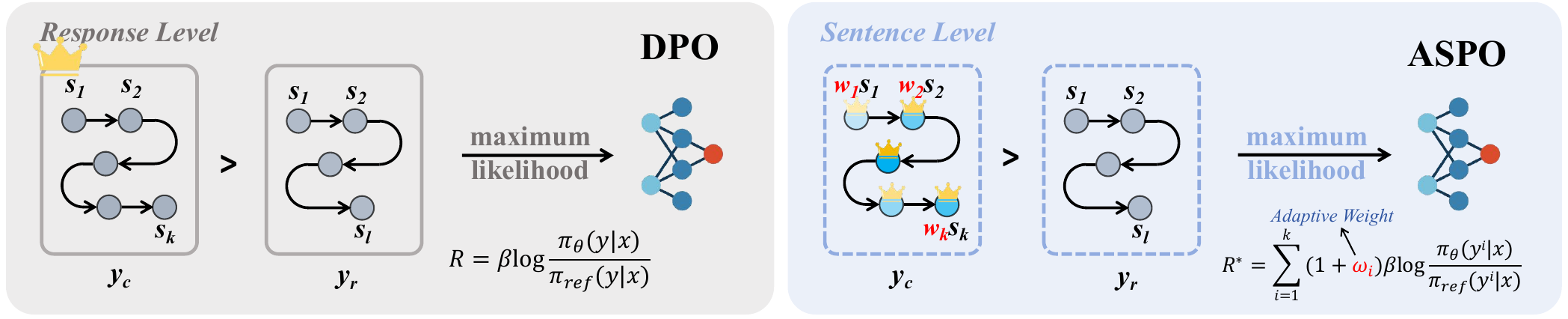}}
\caption{Comparison between DPO and our ASPO. Unlike the traditional DPO algorithm that calculates implicit rewards at the response level, ASPO calculates adaptive rewards using sentences as the smallest units to achieve a fine-grained self-supervised preference optimization method.}
\label{teaserfigure}
\end{center}
\vskip -0.2in
\end{figure*}

Despite these advancements, SFT often leads to hallucinated outputs, causing performance saturation~\cite{zhao2023hadpo, jiang2024hacl}. 
As noted in~\cite{hong2024orpo}, increasing the probability of preferred outputs can inadvertently raise the likelihood of dis-preferred outputs, making models more prone to errors in complex reasoning tasks~\cite{wang2024enhancing, zhang2024improve}. 
Additionally, as response length grows, models become increasingly susceptible to generating hallucinations~\cite{favero2024multi}. 
This issue is exacerbated by machine-generated data containing noisy sentences, where segments may be \textbf{partially} correct or incorrect~\cite{yu2024rlhf-v, lai2024stepdpo}. 
Training on such data risks convergence to suboptimal solutions, emphasizing the need for strategies to suppress undesirable outputs~\cite{liao2024tpo}.

In response, researchers have developed various preference fine-tuning methods to better align MLLMs with human preferences~\cite{sun2024salmon, yuan2024self-rewarding, xie2024monte}. 
Among these, DPO-based methods~\cite{rafailov2024dpo} have gained prominence due to their simplicity and low computational cost~\cite{wang2024mdpo, zhu2024seva}. 
However, traditional DPO methods rely on binary preference data and a coarse reward mechanism, which lacks the fine-grained preference granularity needed to identify specific errors in responses~\cite{liao2024tpo}. 
This limitation hampers the model's ability to refine its reasoning capabilities~\cite{lai2024stepdpo, liao2024tpo}. 
To this end, we introduce a fine-grained preference mechanism that assigns variable reward values, enabling more fine-grained reasoning optimization.

As shown in \cref{teaserfigure}, we present an Adaptive Sentence-level Preference Optimization (ASPO) approach, where each sentence in a response serves as the fundamental unit of preference optimization. 
Unlike traditional DPO, which evaluates preferences at the response level~\cite{li2023silkie, wang2024mdpo}, ASPO systematically evaluates the correctness and importance of each sentence and assigns appropriate rewards. 
By prioritizing accurate reasoning and rejecting errors based on fine-grained criteria, our method efficiently identifies erroneous sentences for targeted optimization. 
This reduces the impact of noisy data and substantially improves the model's comprehensive capability.

Our contributions are summarized as follows:
\begin{itemize}
\item We propose a fine-grained ASPO approach that uses sentences as the fundamental units for preference optimization, surpassing traditional response-level approaches.
\item We design an adaptive reward mechanism that dynamically adjusts rewards at sentence-level, enabling precise preference optimization.
\item We demonstrate significant improvements through extensive experiments and establish the broad applicability of our approach across various multimodal reasoning tasks.
\end{itemize}

\section{Related Work}

\subsection{Multimodal Large Language Models}
With the unprecedented success of LLMs~\cite{touvron2023llama, zhang2022opt}, recent research focused on integrating visual encoders with advanced LLMs to construct MLLMs~\cite{liu2024llava, instructblip}. 
Experimental evidence indicates that MLLM performs effectively across various multimodal tasks~\cite{anil2023gemini, achiam2023gpt4}.

Despite significant advancements in understanding high-level semantics, MLLMs continue to encounter challenges in addressing fine-grained details~\cite{yuan2024osprey, guo2024regiongpt}. 
In complex visual scenarios, these models often struggle to capture critical aspects such as object relationships, attribute alignment, and recognition of local features~\cite{yuan2024osprey, guo2024regiongpt, ouali2025clipdpo}.
Moreover, in tasks like visual question answering or image captioning, models frequently generate hallucinated outputs~\cite{chen2024halc, jiang2024hacl}. 
Existing hallucination problems in models often introduce noise into the generated data, increasing the risk that models trained on these data will converge to suboptimal solutions.

To address these challenges, we propose ASPO that incorporates an adaptive reward mechanism to mitigates the impact of noisy data. 
This mechanism dynamically adjusts reward weights based on the accuracy of response segments, facilitating precise fine-grained optimization.

\subsection{Preference Alignment}
Preference alignment~\cite{chen2023many, ethayarajh2024kto} is widely used to enhance model's instruction-following capabilities~\cite{wang2024sima}. 
Early work primarily relied on Reinforcement Learning from Human Feedback (RLHF)~\cite{leerlaif, wang2023rlhf}.
However, RL-based methods face challenges in stability and efficiency~\cite{rafailov2024dpo}.
DPO~\cite{rafailov2024dpo} addresses these issues by redefining the training objective as a classification loss-based optimization problem. 
By mapping reward functions to optimal policies, the constrained reward maximization problem could be precisely optimized through single-stage policy training~\cite{rafailov2024dpo, wang2024sima}. 

In the multimodal domain, early DPO-based methods relied on response-level rewards~\cite{li2023silkie, zhou2024povid, wang2024sima}. 
For example, SILKIE~\cite{li2023silkie} constructs a vision-language feedback dataset by gathering annotations from multiple MLLMs and employs GPT-4V to evaluate outputs across various dimensions.
MDPO~\cite{wang2024mdpo} addresses the unconditional preference problem by prioritizing image preferences over language preferences.
However, response-level rewards have limitations in granularity and specificity \cite{yoon2024tlcr}, potentially leading to less precise model optimization due to the inclusion of noisy segments in the training data. 
This highlights the need for more fine-grained feedback mechanisms.

Subsequent work has explored much more fine-grained feedback mechanisms at the sentence level \cite{yu2024rlhf-v, leerlaif, ouali2025clipdpo, zhou2024csr}. 
RLHF-V \cite{yu2024rlhf-v} collects human preferences in the form of segment-level corrections on hallucinations and performs dense DPO over corrective human feedback, enhancing the trustworthiness of MLLMs. 
CSR \cite{zhou2024csr} iteratively generates candidate responses, evaluates their rewards, and curates preference data for fine-tuning.

Recent studies have investigated token-level reward mechanisms with notable success~\cite{yoon2024tlcr, cui2024fine, fu2024tldr}. 
FiSAO \cite{cui2024fine} leverages the model’s own visual encoder as a fine-grained verifier to improve vision-language alignment by exploiting token-level feedback from the visual encoder. 
TLDR \cite{fu2024tldr} introduces a token-level detection reward model to provide fine-grained annotations for MLLMs, offering a foundation for post-training methods such as DPO and Proximal Policy Optimization (PPO) \cite{schulman2017ppo}.


Despite these advancements, most approaches rely on model-generated datasets, which tend to be noisy due to hallucinations \cite{yu2024rlhf-v, li2023silkie}. 
Traditional DPO treats all segments of a response equally regardless of their correctness, introducing noise and hindering model optimization.
In addition, most previous methods rely on paid APIs or deploy other external MLLMs, which is not cost-friendly.

To address these issues, we propose ASPO, which dynamically adjusts reward weights of response segments without relying on paid APIs, additional training data or deploying other external MLLMs.
This approach enables more fine-grained preference optimization, alleviating the impact of noise and providing a promising direction for advancing preference alignment research.

\section{Methodology}

In this section, we introduce ASPO for fine-grained multimodal reasoning. 
ASPO incorporates two adaptive reward weights into DPO, i.e., the image-text similarity weight, which optimizes multimodal alignment to mitigate hallucination, and the textual perplexity weight, which enhances text confidence to improve the model's reasoning capabilities.

\subsection{Preliminaries}
RLHF leverages human feedback to train models in scenarios where direct supervision, such as ground-truth labels, is unavailable or insufficient. 
It is widely used to optimize LLMs and MLLMs, aligning their outputs with human preferences.

In RLHF, a Bradley-Terry (BT) reward model is commonly used, defining the human preference distribution as:
\begin{equation}
    p^*(y_{c} \succ y_{r} |x) = \frac{\exp{(r^*(x, y_c))}}{\exp{(r^*(x, y_c))} + \exp{(r^*(x, y_r))}}\,
    \label{eq:bt}
\end{equation}
where $y_c$ and $y_r$ denote the \textit{chosen} and \textit{rejected} responses conditioned on the model's input prompt $x$, and $r^*(x, y)$ represents the latent reward model.

The reward model $r_{\phi}(\cdot)$, parameterized by $\phi$, is trained via maximum likelihood estimation using a preference database $\mathcal{D}$, which is annotated by human evaluators~\cite{yu2024rlhf-v} or advanced AI models~\cite{leerlaif}:
\begin{equation}
    \mathcal{D} = \{x^i, y_c^i, y_r^i\}_{i=1}^N
\end{equation}
The optimization objective maximizes the preference policy:
\begin{equation}
    \max_{\theta}\mathbb{E}_{x,y}\left\{ r_{\phi}(x, y) - \beta \mathbb{D}_\text{KL}[\pi_{\theta}(y|x) | \pi_{\text{ref}}(y|x)] \right\}
    \label{eq:rlhf}
\end{equation}
where the reference model $\pi_\text{ref}$ is typically initialized using the supervised fine-tuned model $\pi_\text{SFT}$ to ensure that the learned parameter $\theta$ does not deviate excessively.
The resulting learned policy $\pi_{\theta}(\cdot)$ aligns more effectively with human intentions.

DPO simplifies RLHF by deriving a closed-form solution~\cite{rafailov2024dpo} from Eq.~\ref{eq:rlhf}, representing the optimal reward model $r^*$ in terms of the learned optimal preference model $\pi^*$ as:
\begin{equation}
r^*(x, y) = \beta \log\frac{\pi^*(y|x)}{\pi_{\text{ref}}(y|x)} + \Delta
\label{eq:form}
\end{equation}
where $\Delta$ is a constant factor. 
Substituting $r^*$ in Eq.~\ref{eq:bt} with Eq.~\ref{eq:form}, the final optimized loss function for DPO is expressed as~\cite{rafailov2024dpo}:
{\small
\begin{equation}
    \mathcal{L}_{\text{DPO}} = -\mathbb{E}_{\mathcal{D}}\left[\log \sigma\left(\beta \log\frac{\pi_{\theta}(y_c|x)}{\pi_{\text{ref}}(y_c|x)} - \beta \log\frac{\pi_{\theta}(y_r|x)}{\pi_{\text{ref}}(y_r|x)}\right)\right]
    \label{eq:dpo}
\end{equation}}
where $\theta$ denotes the trainable parameters in the policy model, $\sigma$ is the logistic function and the $\beta \log\frac{\pi_{\theta}(y|x)}{\pi_{\text{ref}}(y|x)}$ item
can be regarded as an “implicit reward”.
The objective of DPO is to align the  ``implicit reward'' directly with human preference data.

\subsection{ASPO}
ASPO is designed to address the limitations of traditional DPO approache, which optimizes entire responses using binary rewards and fail to account for variations among individual response segments. 
ASPO introduces a sentence-level adaptive reward mechanism that utilizes \textbf{image-text similarity} and \textbf{textual perplexity} metrics to compute fine-grained rewards for each response segment. 
This enables targeted and differentiated optimization, significantly improving the model’s performance.
The overall process is shown in~\cref{alg:ASPO}.

\subsubsection{Rewards of ASPO}
Instead of treating a response as an indivisible unit, ASPO decomposes it into sentences for more precise analysis.
Two critical features are computed for each sentence as follows. 

\textbf{(1) Image-Text Similarity.} 
To ensure the model's output is consistently aligned with the input image, reducing hallucinations and improving model credibility, we measure the semantic relevance of each sentence to the input image as part of the adaptive reward weight.
    
Let $ s_i $ denotes the $i$-th sentence in a response, and let $ x $ represents the corresponding image. 
The similarity function between $ s_i $ and $ x $ is denoted as:
\begin{equation}
    S_{i} = cosine (s_i, x), \quad i = 1, 2, \dots, n
\end{equation}
where $ n $ is the total number of sentences in the response and the similarity score is calculated by CLIP~\cite{clip}.

Next, we apply min-max normalization to scale each similarity score into the range [0,1]:
\begin{equation}
    S'_{i} = \frac{S_{i} - S_{\min}}{S_{\max} - S_{\min}}, \quad i = 1, 2, \dots, n
\label{eq:normalize}
\end{equation}
where $ S'_{i} $ represents the normalized similarity score for the $i$-th sentence with respect to the image $ x $.
$S_{\min}$ and $S_{\max}$ represent the minimum and maximum similarity scores across all sentences, respectively.

\begin{algorithm}[tb]
\small
\caption{ASPO Training Process}
\label{alg:ASPO}
\begin{algorithmic}
   \STATE {\bfseries Input:} Preference dataset $\mathcal{D} = \{(x^i, y_c^i, y_r^i)\}_{i=1}^N$, \\Reference model $\pi_{\text{ref}}$, Policy model $\pi_{\text{$\theta$}}$
   \STATE {\bfseries Output:} Optimized policy model $\pi_{\text{$\theta$'}}$
   \REPEAT
   \FOR{each $(x, y_c, y_r)$ in $\mathcal{D}$}
       \STATE Split $y_c$ into sentences $\{s_1, s_2, \dots, s_n\}$ 
       \FOR{each sentence $s_i$ in $\{s_1, s_2, \dots, s_n\}$}
           \STATE Compute \textbf{Image-Text Similarity} $S'_i$
           \STATE Compute \textbf{Textual Perplexity} $PPL'_i$
           \STATE Compute \textbf{Sentence Weight} $w_i$
       \ENDFOR
       \STATE Calculate adaptive implicit reward margin $\mathcal{M^{*}}$ by \cref{eq:adaptive-reward-margin}
   \ENDFOR
   \STATE Update policy parameters $\theta'$ using gradient descent
   \UNTIL{Convergence or maximum iterations reached}
\end{algorithmic}
\end{algorithm}

\textbf{(2) Textual Perplexity.} 
Perplexity is used to evaluate the quality of a language model, which reflects the uncertainty of the model about a given text. 
The lower the value, the better the model's prediction of the text, that is, the model believes that the text is more likely to appear.
Prior research has shown that outputs with higher model confidence are more likely to be correct~\cite{kuhn2023semantic, chen2024quantifying}. 
Thus, we incorporate the model's predicted output probabilities as part of the adaptive reward weight computation. 
Fluency, coherence, and confidence are assessed using metrics such as log-likelihood, perplexity, or information entropy.
Unless otherwise specified, we use textual perplexity as the default reward weight.

The perplexity of a text sequence is defined as the geometric mean of the inverse probabilities of each word in the sequence, given the preceding context. 
For a sequence of $ N $ words $ w_1, w_2, \dots, w_N $, the perplexity can be expressed as:
\begin{equation}
    PPL = \biggl( \prod_{j=1}^{N} \frac{1}{P(w_j | w_{<j})} \biggl)^{\frac{1}{N}}
\end{equation}
Since the reward coefficient is calculated in units of sentences in our algorithm, assuming that the $i$-th sentence contains $N$ tokens and the prior $i-1$ sentences contain a total of $M$ tokens, using logarithms for computational stability, the perplexity of the $i$-th sentence can be written as:
\begin{equation}
\label{eq:ppl}
\begin{split}
PPL_{i} = \exp \biggl( -\frac{1}{N} & \sum_{j=M+1}^{M+N} \log P(w_j | x, w_{<j}) \biggr)
\end{split}
\end{equation}
where $ P(w_j | x, w_{<j}) $ is the conditional probability of the $j$-th word given the preceding image $x$ and context $ w_{<j} = w_1, w_2, \dots, w_{j-1} $, and $ \exp $ denotes the exponential function.

Since a higher perplexity indicates a higher uncertainty in the model about a given text, we first negate the perplexity at the sentence level and then perform min-max normalization to get the normalized perplexity $PPL'_{i}$ for the $i$-th sentence follows \cref{eq:normalize}.

\textbf{(3) Final Adaptive Weight.} To fully leverage the advantages of different metrics, we integrate them as follows:
\begin{equation}
w_i = \alpha S'_{i} + (1-\alpha) PPL'_{i}
\label{alpha}
\end{equation}
where $\alpha$ is the metric-weighting factor introduced to balance the contribution of different metrics and $w_i$ is used to adjust the reward of each sentence.
  
\noindent\textbf{Sentence-Level Reward Aggregation.} 
In the standard DPO algorithm, the ``implicit reward'' is shared across all steps, and the output of the model at each step receives the same reward weight. 
Specifically, for the triplet $(x, y_c, y_r)$, the ``implicit reward'' margin is mathematically defined as:
\begin{equation}
\label{eq:reward}
\begin{aligned}
\mathcal{M} = \beta &\log\frac{\pi_\theta(y_c\mid x)}{\pi_{\mathrm{ref}}(y_c\mid x)} 
                - \beta\log\frac{\pi_\theta(y_r\mid x)}{\pi_{\mathrm{ref}}(y_r\mid x)} \\
              = \sum_{i=1}^K \beta &\log \frac{\pi_\theta(s^c_i \mid x)}{\pi_{\mathrm{ref}}(s^c_i \mid x)} 
                - \sum_{i=1}^L \beta \log \frac{\pi_\theta(s^r_i \mid x)}{\pi_{\mathrm{ref}}(s^r_i \mid x)}
\end{aligned}
\end{equation}
To address the limitation of the shared reward weight, which fails to differentiate the importance of individual sentences, ASPO introduces an adaptive reward mechanism to adjust the rewards for each sentence.
Specifically, ASPO uses the adaptive weight $w_i$ to modulate the reward strength of each sentence.
The adaptive $M$ is mathematically reformulated as:
\begin{equation}
\label{eq:adaptive-reward-margin}
\begin{split}
\mathcal{M^{*}} = & \frac{R_c}{{R}^*_c}\sum_{i=1}^K \beta (1+w_i) \log \frac{\pi_\theta(s^c_i \mid x)}{\pi_{\mathrm{ref}}(s^c_i \mid x)} \\
& - \beta\log\frac{\pi_\theta(y_r\mid x)}{\pi_{\mathrm{ref}}(y_r\mid x)}
\end{split}
\end{equation}
where ${R_c}$ and ${R}^*_c$ represent the original and the reweighted sum of the sentence-level implicit rewards of chosen response, respectively.
We introduce the $\frac{R_c}{{R}^*_c}$ item to normalize the influence of sentence-level rewards across responses of varying lengths, which prevents longer responses from disproportionately benefiting from a total weight increase, aligning better with our objective of implementing fine-grained rewards.

Note that when a response contains only one sentence, the adaptive weight $w_i$ is normalized to 0 and our method degenerates into the standard DPO.

\begin{table*}
	\centering
        \setlength{\tabcolsep}{1.2pt}
	\begin{tabular}{lccccccccccc}
		\toprule[1pt]
        Method & MMVet  & MMB$^\text{D}$ & MMB$^\text{T}$ & MMB$^\text{C}$ & SEED$^{I}$ & LLaVA$^\text{W}$ & SQA$^{I}$ & GQA & POPE & SHR $(\downarrow)$ & Avg. \\
		\midrule[1pt]   
        BLIP-2-7B & 22.4 & -- & -- & -- & 46.4 & 38.1 & 61.0 & 41.0 & 85.3 & -- & -- \\
        MiniGPT-4-7B & 22.1 & 23.0 & -- & -- & 42.8 & -- & -- & -- & 74.5  & -- & -- \\
        Shikra-13B & -- & 58.8 & -- & -- & -- & -- & -- & -- & -- & --  & -- \\
        LLaVA-7B & 26.7 & 34.1 & -- & 14.1 & 25.5 & 63.0 & 38.5  & -- & -- & -- & -- \\
        IDEFICS-7B  & --  & 48.2 & -- & 25.2 & -- & -- & -- & 38.4 & -- &  -- & -- \\
        IDEFICS-65B  & -- & 54.5 & -- & 38.1 & -- & -- & -- & 45.2 & -- &  -- & -- \\
        mPLUG-Owl2-7B & 36.2 & 64.5 & -- & -- & -- & -- & 68.7 & 56.1 & 85.8 & -- & -- \\
        Qwen-VL-7B & -- & 38.2 & -- & 7.4 & 56.3 & -- & 67.1 & 59.3 & --  & -- & -- \\
        Qwen-VL-chat-7B & --  & 60.6 & -- & 56.7 & 58.2 & -- & 68.2 & 57.5 & --  & -- & -- \\
        \midrule
        InstructBLIP-13B & 25.6  & 33.5 & 34.0 & \textbf{26.3} & 45.4 & 58.2 & 44.7 & 48.1 & 78.9 & 51.2 & 43.86\\
        +DPO  & 26.1  & 33.5 & 34.3 & 26.1 & 45.8 & 62.7 & 44.3 & 47.5 & 80.2  & 53.8 & 44.50 \\ 
        \rowcolor{gray!10}
        \textbf{+ASPO}  & \textbf{27.0} & \textbf{33.7} & \textbf{35.1} & 26.0 & \textbf{46.3} & \textbf{67.4} & \textbf{45.1} & \textbf{48.2} & \textbf{83.8} & \textbf{50.5} & \textbf{45.84} \\
        LLaVA-1.5-7B  & 30.5  & 64.3 & 66.4 & 58.3 & 65.7 & 63.4 & 66.8 & \textbf{62.0} & 85.9  & 36.7 & 62.59 \\
        +DPO  & 33.3  & 64.7 & 66.4 & 58.3 & 66.1 & 65.7 & 66.4 & 61.0 & 86.2  & 40.1 & 63.12 \\
        \rowcolor{gray!10}
        \textbf{+ASPO} & \textbf{35.3} & \textbf{65.6} & \textbf{67.7} & \textbf{59.5} & \textbf{66.3} &\textbf{75.7} & \textbf{67.7} & \textbf{62.0} & \textbf{86.6}  & \textbf{33.9} & \textbf{65.16} \\
        LLaVA-1.5-13B  & 35.4 & 67.7 & 67.0 & 63.6 & 68.2 & 70.7 & 71.6 & 63.3 & 85.9 & 37.2 & 65.93 \\
        +DPO  & 38.8  & 69.6 & 69.9 & 64.1 & 68.2 & 74.7 & 70.2 & 62.3 & 85.5  & 47.4 & 67.03 \\ 
        \rowcolor{gray!10}
        \textbf{+ASPO}  & \textbf{41.2} & \textbf{70.4} & \textbf{70.7} & \textbf{64.7} & \textbf{68.5} & \textbf{82.0} & \textbf{71.8} & \textbf{63.4} & \textbf{86.5} & \textbf{34.8} & \textbf{68.80} \\
        \bottomrule[1pt]
	\end{tabular}
\caption{Comparison results with state-of-the-art methods.  
ASPO consistently improves base model's performance on multiple benchmarks. The last column shows the average values of all metrics except for SHR.}
\label{tab:compare-with-sota}
\end{table*}


\begin{table*}[!t]
\centering
\begin{tabular}{lcccccccc}
\toprule
Method & Level & MMVet & MMB$^\text{D}$ & LLaVA$^{\mathrm{W}}$ & SQA$^{\mathrm{I}}$ & POPE & Avg. \\
\midrule
LLaVA-1.5-7B & -- & 30.5 & 64.3 & 63.4 & 66.8 &85.9 & 62.18 \\
+ Vlfeedback & Response & 31.2 & 64.0 & 62.1 & 66.2 &83.7 & 61.44 \\
+ Human-Prefer & Response & 31.1 & 63.4 & 63.7 & 65.8 &81.5 & 61.10 \\
+ POVID & Response & 31.8 & 64.9 & 68.7 & 68.8 & \textbf{86.9} & 64.22 \\
+ SIMA & Response & 31.6 & 64.9 & 66.1 & 69.1 & 86.5 & 63.64 \\
+ HA-DPO & Response & 30.5 & 64.4 & 64.2 & 68.5 &85.8 & 62.68 \\
+ RLHF-V & Sentence & 30.9 & 63.6 & 65.4 & 67.1 &86.2 & 62.64 \\
+ RLAIF-V & Sentence & 30.5 & 63.4 & -- & 68.4 &81.5 & -- \\
+ CLIP-DPO & Sentence & -- & 64.9 & -- & 67.6 &85.8 & -- \\
+ CSR iter-1 & Sentence & 32.2 & 64.7 & 69.7 & 70.3 & 85.8 & 64.54 \\
+ CSR iter-3 & Sentence & 33.9 & 65.4 & 71.1 & \textbf{70.7} & 85.9 & 65.40 \\
+ FiSAO & Token & 30.7 & 64.8 & -- & 69.3 & 85.7 & -- \\
\rowcolor{gray!10}
+ \textbf{ASPO} &\textbf{Sentence} & \textbf{35.3}  & \textbf{65.6} & \textbf{75.7} & 67.7 & 86.6  & \textbf{66.18} \\
\bottomrule
\end{tabular}
\caption{Comparison results of different preference optimization-based methods. 
Most baseline results are from \cite{zhou2024povid}.
Benchmarks like GQA in \cref{tab:compare-with-sota} are not included since most baseline methods have not been tested on them.}
\label{tab:benchmark_dpo}
  \vspace{-2mm}
\end{table*}

\begin{table}
  \begin{minipage}{0.99\linewidth}
\centering
\scalebox{0.80}
{
\begin{tabular}{p{1.0cm} p{7.0cm} }
\toprule
 \multicolumn{2}{l}{\bf Sample of Visual Qustion Answering:}  \\
\midrule
&  \includegraphics[height=3.5cm]{vqa.png} \\
User & Are the trees taller than the giraffes? \\
\midrule
LLaVA-1.5 & \textcolor{red}{Yes, the trees are taller than the giraffes, as they are reaching up to eat leaves from the trees.} \\ 
\midrule
ASPO (Ours) & No, the trees are not taller than the giraffes. The giraffes are eating leaves from the trees, which are at a height that is accessible to them. \\
\bottomrule
\end{tabular}
}
\captionof{table}{Our approach is more faithful to the image input and demonstrates stronger reasoning capabilities.
Hallucination outputs are shown in \textcolor{red}{red}. }
\label{tab:vqa case}  
  \end{minipage}
  \vspace{-2mm}
\end{table}

\subsubsection{Optimization Objective of ASPO}
The ASPO loss function extends the traditional DPO objective by incorporating sentence-level granularity. 
For a preference pair $(y_c, y_r)$, the optimization objective becomes:
\begin{equation}
\mathcal{L}_{\text{ASPO}} = -\mathbb{E}_{\mathcal{D}}\left(\log \sigma\mathcal{M^{*}}\right)
\end{equation}
By focusing on individual sentences, ASPO allows the model to refine specific response segments, leading to improved fine-grained understanding. 
Adaptive weighting enables the model to prioritize high-quality or critical sentences while minimizing the influence of less relevant ones. 
ASPO can be seamlessly integrated into existing DPO-based training pipelines with minimal adjustments.


\section{Experiment}
In this section, we conduct comprehensive experiments to evaluate the effectiveness of the proposed ASPO. 
We detail the experimental setups, results, and analysis in the following sections.

\subsection{Experimental Setups}

\noindent\textbf{Pretrained Models.} 
To assess ASPO's generalization, we test it on the popular open source MLLMs in different architectures and sizes.
LLaVA-v1.5-7B and LLaVA-v1.5-13B~\cite{liu2024llava1.5} uses MLP as an alignment module, while InstructBLIP-13B~\cite{instructblip} uses Q-Former~\cite{li2023blip2}, a query-based alignment module to align multimodal features.
We use the models after SFT as base models and conduct DPO training on them.

\begin{figure}[ht]
\begin{center}
\centerline{\includegraphics[width=0.5\textwidth]{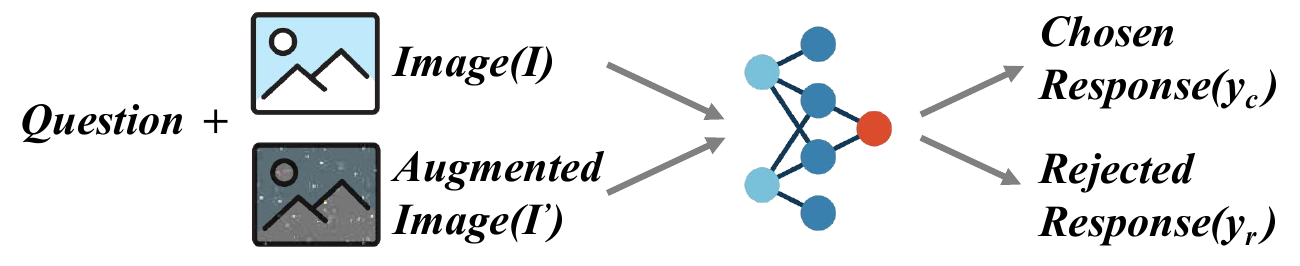}}
\caption{Data collection pipeline.}
\label{fig:data}
\end{center}
\vskip -0.2in
\end{figure}

\noindent\textbf{Preference data.} 
We sample about 20K instructions from LLaVA-Instruct-150K~\cite{liu2024llava} and go through SeVa pipline~\cite{zhu2024seva} to produce the preference dataset.
The data collection pipeline is shown in~\cref{fig:data}.
Specifically, diffusion noise is added to the images in the instruction data, and the original and augmented images with the instruction are then fed into the MLLM to be train to obtain the chosen and rejected responses, respectively. 
The identical paired preference data is filtered out and we finally get about 16K preference data pairs. 
For all experiments, the noise steps are set to 500.
In this way, preference data is obtained with the same style and distribution as previous studies~\cite{zhao2023hadpo} suggesting that unifying factors such as style and distribution in preference data could enhance model optimization.

\begin{table}[t]
  \begin{minipage}{0.99\linewidth}
\centering
\scalebox{0.80}
{
\begin{tabular}{p{1.0cm} p{7.0cm} }
\toprule
 \multicolumn{2}{l}{\bf Sample of Image Captioning:}  \\
\midrule
&  \includegraphics[height=3.5cm]{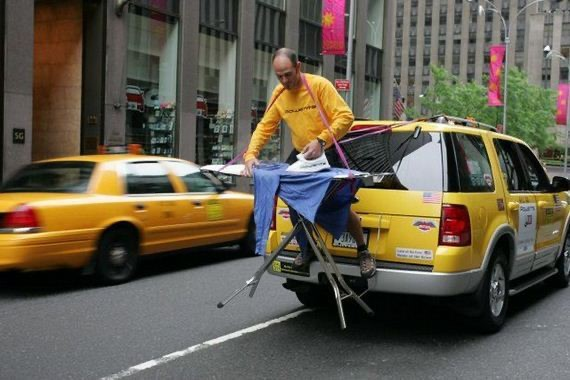} \\
User & Describe this photo in detail. \\
\midrule
LLaVA-1.5 & The image features a man standing on the back of a yellow SUV, ironing a shirt. He is using an ironing board placed on the back of the vehicle, which is an unusual and creative way to iron clothes while on the go. \textcolor{red}{In the background, there are two other people visible, one on the left side and another on the right side of the scene.} \textcolor{red}{A potted plant} can be seen in the middle of the scene, adding a touch of greenery to the urban environment. \textcolor{red}{Additionally, there is a traffic light in the background}, indicating that the scene takes place in a busy city area. \\ 
\midrule
ASPO (Ours) & The image features a man ironing a shirt while standing on the back of a yellow SUV. It is driving down a city street, and the man appears to be multitasking by attending to his clothes and navigating the busy street. There is also a yellow taxi on the street. Additionally, some trees can be seen on the sidewalk, adding a touch of greenery to the urban environment. \\
\bottomrule
\end{tabular}
}
\captionof{table}{ASPO provides more reliable and contextually accurate descriptions.
Hallucination outputs are shown in \textcolor{red}{red}. }
\label{tab:caption case}  
  \end{minipage}
  \vspace{-2mm}
\end{table}

\begin{table*}
	\centering
        \setlength{\tabcolsep}{1.6pt}
	\begin{tabular}{lccccccccccc}
		\toprule[1pt]
        Method & MMVet  & MMB$^\text{D}$ & MMB$^\text{T}$ & MMB$^\text{C}$ & SEED$^{I}$ & LLaVA$^\text{W}$ & SQA$^{I}$ & GQA & POPE & SHR $(\downarrow)$ & Avg. \\
        \midrule
        LLaVA-1.5-7B  & 30.5  & 64.3 & 66.4 & 58.3 & 66.1 & 63.4 & 66.8 & \textbf{62.0} & 85.9  & 37.9 & 62.59 \\
        +DPO  & 33.3  & 64.7 & 66.4 & 58.3 & 66.1 & 65.7 & 66.4 & 61.0 & 86.2  & 40.1 & 63.12 \\
        \textbf{+ASPO-\textit{S}} & \textbf{35.6} &65.5 &67.3 &59.5 &66.1 & 74.6 &67.4 & 61.7 &86.5  & 34.6 & 64.91 \\
        \textbf{+ASPO-\textit{P}} &34.9 & \textbf{65.6} & \textbf{67.9} & \textbf{59.8} & \textbf{66.4} & 75.4 & \textbf{67.9} & 61.9 & \textbf{86.6}  & 35.2 & \textbf{65.16} \\
        \rowcolor{gray!10}
        \textbf{+ASPO-\textit{S+P}} &35.3 & \textbf{65.6} & 67.7 & 59.5 & 66.3 & \textbf{75.7} & 67.7 & \textbf{62.0} & \textbf{86.6}  & \textbf{33.9} & \textbf{65.16} \\
        \bottomrule[1pt]
	\end{tabular}
\caption{Ablation study of reward weight.
\textit{S} represents the Image-Text Similarity and \textit{P} represents Textual Perplexity.
The last column shows the average values of all metrics except for SHR.}
\label{tab:ablation_reward}
\end{table*}

\begin{table*}
	\centering
        \setlength{\tabcolsep}{1.1pt}
	\begin{tabular}{lcccccccccccc}
		\toprule[1pt]
        Method & Level & MMVet  & MMB$^\text{D}$ & MMB$^\text{T}$ & MMB$^\text{C}$ & SEED$^{I}$ & LLaVA$^\text{W}$ & SQA$^{I}$ & GQA & POPE & SHR $(\downarrow)$ & Avg. \\
        \midrule
        DPO & Response  & 33.3  & 64.7 & 66.4 & 58.3 & 66.1 & 65.7 & 66.4 & 61.0 & 86.2  & 40.1 & 63.12 \\
        ASPO & Response  & 20.9  & 58.6 & 59.8 & 50.7 & 58.2 & 58.7 & 63.3 & 50.7 & 73.2  & 72.1 & 54.90 \\
        \rowcolor{gray!10}
        \textbf{ASPO} & \textbf{Sentence} &\textbf{35.3} & \textbf{65.6} & \textbf{67.7} & \textbf{59.5} & \textbf{66.3} & \textbf{75.7} & \textbf{67.7} & \textbf{62.0} & \textbf{86.6}  & \textbf{33.9} & \textbf{65.16} \\
        ASPO & Token  & 30.3  & 61.3 & 62.3 & 54.8 & 61.8 & 62.1 & 66.6 & 59.1 & 77.7 & 59.1 & 59.56 \\
        \bottomrule[1pt]
	\end{tabular}
\caption{Ablation study of reward level on LLaVA-1.5-7B.
We calculate adaptive rewards at the response level, sentence level, and token level to explore the most appropriate solution.
The last column shows the average values of all metrics except for SHR.}
\label{tab:ablation_reward_level}
  \vspace{-2mm}
\end{table*}

\noindent\textbf{Baselines.} 
We compare ASPO with several previous State-Of-The-Art (SOTA) open-source MLLMs and preference optimization-based methods.
The \textbf{open-source MLLMs} include BLIP-2~\cite{li2023blip2}, MiniGPT-4~\cite{zhu2023minigpt4}, InstructBLIP~\cite{instructblip}, LLaVA~\cite{liu2024llava}, LLaVA-v1.5~\cite{liu2024llava1.5}, Shikra~\cite{chen2023shikra}, Qwen-VL~\cite{bai2023qwen}, Qwen-VL-Chat~\cite{bai2023qwen}, mPLUG-Owl2~\cite{ye2024mplug} and IDEFICS~\cite{laurenccon2024obelics}. 
For \textbf{preference optimization}-based methods, we compare ASPO with SILKIE (Vlfeedback)~\cite{li2023silkie}, LLaVA-RLHF (Human-prefer)~\cite{sun2023llava-rlhf}, POVID~\cite{zhou2024povid}, SIMA~\cite{wang2024sima}, HA-DPO~\cite{zhao2023hadpo}, RLHF-V~\cite{yu2024rlhf-v}, RLAIF-V~\cite{leerlaif}, CLIP-DPO~\cite{ouali2025clipdpo}, CSR~\cite{zhou2024csr} and FiSAO~\cite{cui2024fine}.

\noindent\textbf{Evaluation Benchmarks} 
We evaluate the performance of ASPO on several widely used multimodal benchmarks across different dimensions, including comprehensive benchmarks (MMVet~\cite{yu2023mmvet}), MMBench-Dev~\cite{liu2025mmbench}, MMBench-Test~\cite{liu2025mmbench}, MMBench-Chinese~\cite{liu2025mmbench}, SEED-Image~\cite{li2023seedbench}, LLaVA-bench-in-the-wild~\cite{liu2024llava}, hallucination benchmarks (POPE~\cite{li2023pope}, SHR~\cite{zhao2023hadpo}) and general VQA benchmarks (ScienceQA-Image~\cite{lu2022sqa}, GQA~\cite{hudson2019gqa}).

\noindent\textbf{Implementation Details}
The hyperparameter settings for both LLaVA-v1.5-7B and LLaVA-v1.5-13B are identical. 
Specifically, both models are trained for 1 epoch using a batch size of 10.
The learning rate and weight decay are set as 2e-6 and 0, respectively with the learning rate adjusted by a cosine scheduler following prior research~\cite{zhu2024seva}. 
We fine-tune all linear layer parameters using LoRA~\cite{hu2021lora}, setting its rank to 1024 and $\alpha$ to 2048.
The scale parameter $\beta$ in ASPO is fixed as 0.1. 
For InstructBLIP-13B, we set LoRA's rank to 64 and $\alpha$ to 16 following prior research~\cite{zhao2023hadpo}. 
$q_{proj}$, $k_{proj}$, $v_{proj}$ in the language model are fine-tuned. 
We employ a learning rate of 4e-6 with a cosine learning rate scheduler and a batch size of 1, while $\beta$ is set as 0.1. 
The metric-weighting factor $\alpha$ in \cref{alpha} is set to 0.5.
All of the experiments are conducted on 8 NVIDIA-L20Z-80GB GPUs for 1 epoch, every training takes less than 1 hour.
ASPO and standard DPO share the same configuration above.

\subsection{Experimental Results}
As shown in~\cref{tab:compare-with-sota}, ASPO improves the performance of the base models across the board, demonstrating the effectiveness of our approach.
From the experimental results, we can see that from InstructBLIP-13B, LLaVA-1.5-7B to LLaVA-1.5-13B, the performance of these models continues to improve and their improvement shows the same trend after training with our method, which is in line with our expectations. 
This further verifies that the credibility of the basic models is positively correlated with their performance. 
\begin{figure}[ht]
\begin{center}
\centerline{\includegraphics[width=0.5\textwidth]{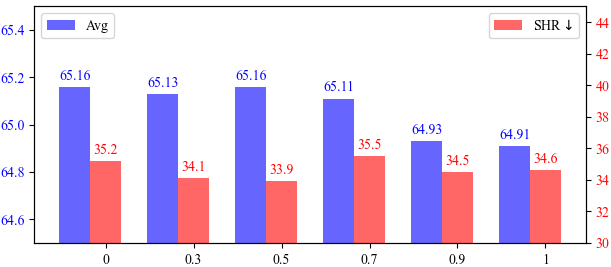}}
\caption{Ablation study of the metric-weighting $\alpha$.}
\label{fig:alpha}
\end{center}
\vskip -0.6in
\end{figure}
Therefore, as the performance of the basic models increases, their credibility becomes higher and higher, and the adaptive rewards calculated based on this become much more accurate, so the improvement in model performance is much larger. 
We show the comparison results of our method with other preference optimization-based methods in~\cref{tab:benchmark_dpo} and ASPO achieves the highest average score of 66.18 across all evaluated datasets, which is notably higher than the baseline LLaVA-1.5-7B model's average score of 62.18. 
This improvement is consistent across most individual datasets as well. 
In addition, compared with reward mechanisms at different levels, our method shows certain advantages.
These results strongly support the effectiveness of our method.
The example results in~\cref{tab:vqa case} show that our method is more faithful to the image input and enhances the model's reasoning capabilities.
Additionally, from the example in~\cref{tab:caption case} on the Image Captioning task, it can be seen that after optimization with our method, the model's output is more concise and contains fewer hallucinations.

\subsection{Ablation Study}
In this section, we conduct comprehensive ablation experiments on the LLaVA-v1.5-7B model to verify the effectiveness of our method. 
As shown in~\cref{tab:ablation_reward}, we first train the model using the standard DPO algorithm, but from the experimental results we only see a slight improvement in model performance, and the performance of the model even decreases on some benchmarks. 
This is mainly due to the fact that a considerable number of preference data pairs have overlapping segments, which introduces a lot of noise into our training process. 
However, our method can deal with this problem well.
We also performed ablation on the level of adaptive rewards to explore the most appropriate reward level. 
The results are shown in~\cref{tab:ablation_reward_level}.
We first tested at the response level. 
Experimental results show that adaptive rewards at the response level significantly reduce model performance. 
This is because the training data itself contains a lot of noise, and the final implicit reward is easily affected by extremely high or low values in the overall response, which affects the stability of model training.
The token-level reward is also inappropriate. 
This is related to the part of the speech of tokens in the response. 
For conjunctions, prepositions, etc., the model often shows higher confidence, while the tokens that really contain key information are not fully optimized. 
We will further study this issue in future work.
In contrast, sentence-level adaptive rewards achieve the best performance.
\cref{fig:alpha} shows the ablation results of the metric-weighting factor $\alpha$. 
We present the changes in both SHR and the average values of all other benchmarks excluding SHR. 
Overall, our method achieves the best results when $\alpha$ is set to 0.5.

\section{Conclusion}

In this paper, we introduce ASPO, a method designed to overcome the limitations of traditional DPO approaches, which treat entire responses as indivisible units during optimization. 
By employing a fine-grained optimization strategy that treats sentences as the fundamental units, ASPO enables more precise and effective alignment of MLLMs with human preferences.
By moving beyond response-level optimization to embrace a sentence-level approach, our method opens new avenues for improving the comprehensive capabilities of MLLMs, paving the way for more accurate and reliable alignment with human preferences.

\section{Limitations}
\label{sec:limitation}
Despite these advances, there are still areas for further exploration. 
One limitation of our method is the reliance on pre-defined metrics, such as image-text similarity and textual perplexity, which may not fully capture the complexity of human preferences in certain scenarios. 
Future work could explore incorporating more sophisticated metrics or leveraging reinforcement learning to refine the reward mechanism further. 
Additionally, the potential impact of ASPO on other aspects, such as robustness and generalization, warrants further investigation.
Finally, due to computational resource limitations, we did not extend the experiments to the base models with much more parameters to provide a clear picture of the scalability of ASPO.

\section*{Acknowledgments}
This work was supported in part by the National Natural Science Foundation of China under Grants 62372380 and U22B2036, Key Research and Development Program of Shaanxi Province under Grants 2024GX-ZDCYL-01-05, the Natural Science Basic Research Program of Shaanxi under Grants 2024JC-YBMS-513, Key Research and Development Program of Zhejiang Province under Grants 2024C01025, and Key Research and Development Program of Hangzhou under Grants 2024SZD1A23.


\bibliography{custom}



\end{document}